\RequirePackage{fix-cm}
\RequirePackage{etex} 

\documentclass[twocolumn]{svjour3}          % twocolumn
\smartqed  % flush right qed marks, e.g. at end of proof

\usepackage{graphicx}
\usepackage{epstopdf}
\usepackage{epsfig,graphics}
\usepackage{latexsym}
\usepackage{amsfonts}
\usepackage{amssymb}
\usepackage{multirow}
\usepackage{booktabs}
\usepackage{cite}
\usepackage{citeref}
\usepackage{subfigure}
\usepackage{amsmath}
\usepackage[misc]{ifsym}
\usepackage[export]{adjustbox}
\usepackage[T1]{fontenc}
\usepackage{hyperref}
\usepackage{mathtools,xparse}
\usepackage{pgfplots}
\usepackage{tikz}
\usepackage{tkz-euclide}
\usetkzobj{all}
\usetikzlibrary{positioning}
\usetikzlibrary{shapes,arrows}
\usetikzlibrary{shadows, positioning,arrows.meta,shapes.geometric}
\usepackage{wrapfig}
\usepackage{multirow}
\usepackage{textcomp}

\begin{document}

\title{On improving learning capability of ELM and an application to brain-computer interface}%\thanks{Grants or other notes



\author{Apdullah Yay{\i}k \and Yakup Kutlu \and G\"{o}khan Altan}


\institute{A. Yay{\i}k (\Letter)  \at Military Academy, \\National Defence University, Ankara, Turkey\\  email: apdullahyayik@gmail.com\\
Y. Kutlu \at Department of Computer Engineering,\\\.{I}skenderun Technical University, Hatay, Turkey\\
G. Altan \at Department of Computer Engineering,\\ Mustafa Kemal University, Hatay, Turkey
}

\date{Received: date / Accepted: date}

\maketitle
\begin{abstract}
As a type of pseudoinverse learning, extreme learning machine (ELM) is able to achieve high performances in a rapid pace on benchmark datasets. However, when it is applied to real life large data, decline related to low-convergence of singular value decomposition (SVD) method occurs. Our study aims to resolve this issue via replacing SVD with theoretically and empirically much efficient 5 number of methods: lower upper triangularization, Hessenberg decomposition, Schur decomposition, modified Gram Schmidt algorithm and Householder reflection.

Comparisons were made on electroencephalography based brain-computer interface classification problem to decide which method is the most useful. 
Results of subject-based classifications suggested that if priority was given to training pace, Hessenberg decomposition method, whereas if priority was given to performances Householder reflection method should be preferred. 


\end{abstract}

\section{Introduction}
\label{intro}
Extreme learning machine (ELM) $-$type of pseudoinverse learning (PIL)$-$ has attracted so much interest from various disciplines in that it is able to successfully solve non-linear pattern recognition problems. %Such as, electrocardiography (ECG) arrhythmia classification \cite{r47}, face recognition \cite{r48, r49}, handwritten character recognition \cite{r52}, epileptic seizure detection \cite{r50} and sales forecasting \cite{r51}. 

Over the last decade, several approaches that improved its learning and generalization capability using orthogonalization \cite{r13, ref58, r31, ref51}, neuron number optimization \cite{r3, r4, r5, r9, ref59, ref70}, selecting effective input weights \cite{r6}, bidirectional learning that considers network error \cite{r8} have been proposed.

Random vector functional link (RVFL) net is another special type of PIL. However, it has not attracted interests as much as ELM has, although being introduced almost 20 years before. In its structure, enhancement nodes are placed between input and output layer (this idea represents classical hidden layer), and input nodes have direct link with output layer \cite{refx2}. Schimdt et al  \cite{refx1} noted that fixing hidden weights to uniformly random values between $-1$ and $+1$ and computing output weighs with fisher solution (using the same idea in ELM that can be evaluated as an inspiring approach for ELM developer) was much more efficient than classical backpropagation (BP) algorithm. They showed redundancy of learning hidden layer for pattern recognition. Pao et al  \cite{refx2} compared the performances between RVFL net and BP algorithm and reached that RVFL outperforms BP. 
%In 1994, Verma and Mulawka [3] introduced learning algorithm based on pseudoinversing, however it has unnecessary parts that makes it also inefficient. In the output layer logarithmic sigmoid activation function was used and pseudoinversing with QR decomposition was performed after throwing it away. 
Guo et al \cite{refx4}, suggested setting number of hidden nodes equal to number of input nodes to avoid rank deficiency. In recent studies, researchers observed critical open problems needed to be focused on. In 2015, Zhang and Suganthan \cite{refx7} remarked that a robust method should be investigated to handle rank deficiency. 
Our previous study \cite{ref59} together with current research sheds light to deal with rank deficiency  difficulties mentioned in Zhang and Suganthan \cite{refx7}. Zhang and Suganthan \cite{refx11} did not explain how to optimize regularization parameter and the preferred pseudoinverse method that complies with Moore-Pensore conditions. In 2018, Guo  \cite{refx9} proposed methods to optimize hidden layer neuron number for PIL systems with considering overfitting together with underfitting. Zhang and Suganthan \cite{refx11} approved that direct input-output connection has significant effect on performance. In addition, they proposed that regularized least square solution for learning procedure outperforms Moore-Pensore pseudoinverse solution. Zhang and Suganthan \cite{refx12} used RVFL net in the last layer of convolutional neural networks $-$fully-connected layer$-$ instead of BP method and achieved promising performance in visual tracking application. Ren et al \cite{refx10} compared randomized neural networks with (RVFL nets) and without direct input-output connections and reached that RVFL nets outperforms in several sample datasets. In addition, they contributed an enhancement node optimization method based on time series cross validation. In current study structure of proposed ELM model and RVFL net in terms of theoretical background, parameter regularization for pseudoinverse, pitfalls and open problems were explained.
 
In BP algorithm, minimizing loss function is controlled using a different data, namely validation set, excluded from training set to avoid overfitting, in other words to make it gain generalization capability. However, ELM cannot gain a generalization capability using necessarily the same approach in BP. Recently, Cao et al. \cite{ref72} proposed that ELM with SVD method can gain generalization ability with considering performance of pseudoinversing in leave-one out (LOO) model when computing loss function $-$mean square error (MSE). Therefore, they put $1-$diag($\mathbf{H}\mathbf{H}^+$) regularization term to the denominator of classical MSE, in order to observe the effect of pseudoinversing. However, this approach ignores a basic problem that is when $\mathbf{H}\mathbf{H}^+$ is positive definite matrix, divide by zero error occurs. Therefore proposed method should be numerically stabilized. Furthermore, elements at above and below the diagonal of $\mathbf{H}\mathbf{H}^+$  should converge to zero, however only convergence to zero at diagonal was considered in regularization term.

Computational stability of ELM hasn't been proven, yet. Computing weights of hidden layer outputs from $\mathbf{H}w=t$ when $\mathbf{H}$ is not full column rank or ill-conditioned requires stabilized algorithm that can work in large datasets. In conventional ELM, this issue is solved using SVD \cite{r42}. Unfortunately, SVD has disadvantages, such as being slow and low convergence to real solution. Our study addressed this issue by proposing to replace SVD with 5 number of efficient methods; lower upper (LU) triangularization, Hessenberg decomposition, Schur decomposition, modified Gram Schmidt (MGS) process and Householder reflection. 

EEG data don't let researchers to draw inferences about how and when the brain responds to a specific stimuli. Researchers started to carry out measuring event related potentials (ERPs) to reach this aim. ERPs are gained using overlapping EEG time series across channels from onset of stimuli to a specific timing and grand averaging. The reason behind performing averaging is that each trial includes both signal and noise. Because the signal is necessarily almost similar, whereas noise varies across trials, averaging provides removing those unwanted noise while signal is slightly effected. If the experimental analysis is performed to reveal differences between two conditions without considering electrophysiological dynamics, ERP is the fastest and easiest method with high temporal precision and accuracy \cite{ref71}.

P300 (P3) wave is an ERP endogenous component elicited in the course of decision making process. It is identified as a positive deflection about 250 to 500 ms after the stimuli onset \cite{r14}. Latency when elicited by a visual stimuli is 350 to 450 ms, while latency elicited by an auditory or a tactile stimuli is 150 to 300 ms \cite{r24}. To elicit P3 wave, the most common and validated method is implementing the oddball paradigm, in which low-probability target items are mixed with high-probability non-target items \cite{r23}.  

P3 wave has proven to be a trusty response, to control brain-computer interface (BCI) for neurologically disordered people \cite{r32}. It was used in; mouse control \cite{r25}, spelling \cite{r26, r27, r28}, robotic arm control \cite{r29}, visual object detection \cite{r30, r31, r46}. Additionally, P3 waves were used for psychological assessments in patients with cognitive and attention disorders such as alzheimer\'s dementia \cite{r17}, schizophrenia \cite{r18}, alcohol dependence \cite{r19}, and speech disorders \cite{r20}. 

In 1988, Farwell and Dolchin introduced the first BCI based spelling system that uses P3 wave of 50 Hz. Row/Column flashing model was used in visual stimulus. The system was tested on 4 subjects who wrote 5 letters. As a classifier stepwise linear discriminant analysis, peak picking, area and covariance were used and compared. This was the pioneering study that uses P3 wave and oddball paradigm for a mental prosthesis \cite{ref63}. 

In 2007, Hoffman et. al introduced a BCI system that could perform classifying  visual objects using P3 wave of 2048 Hz sampled EEG data of 32 channels. The system was tested on 9 subjects. Any feature extraction algorithm was used, instead down-sampling was applied, 1000 ms segments were extracted from data although inter stimulus was 400 ms (600 ms of segments overlap) and for denoising EEG data windsorizing operation of 10\% was performed \cite{r30}.

In 2009, Takano et. al investigated green, blue flicker stimuli effect of visual stimuli on P3 wave from 10 channels. EEG data from 10 able-bodied subjects were recorded and it was reached that impact of green and blue flicker stimulus were higher than that of white and gay flicker ones on P3 wave \cite{ref64}.

In 2010, Donnerer M. and Steed A. investigated in 3-dimensional (3D) controlling system using P3 wave from 8 channel EEG data in 3D environment and proved possibility of developing a BCI. Objects paradigm, tiles paradigm and 36 spheres paradigm were tried to integrate BCI to an augmented reality style control system. 36 spheres is exactly the same paradigm that is known as single character flashing in conventional BCI speller scenario \cite{ref65}.

%In 2012, Kaufmann et. al investigated whether N200 wave of ERP can be used in a BCI based  speller system, or not. They reached that the best way was using only P3 wave or both P3 and N2 waves, but not just N2 wave. 256 Hz sampled EEG signals from 12 channels were used and results were proven with tests on 51 subjects. Important properties of this study are, flash duration (31.25 ms) and inter stimulus duration (125 ms) are very short and filtering range (0.1 Hz. and 30 Hz.) of raw EEG signals is broad. For denoising raw EEG signals any operation was performed except for filtering. It is seen that though high overlapping on ERPs and not-denoising raw EEG signals, successful results were reached \cite{ref66}.

In 2012, Jin et. al proposed BCI based speller system that relied on both P3 wave and motion-onset visual potential. In this combined system visual stimuli effects consisted of colour change (blue/green, white/grey), moving (with 6   speed in 3 cm distance) and both. The system was tested on 10 subjects. EEG data were recorded from 12 channels and sampled at 36.6 Hz. Besides, three different range of duration; 0-800 ms, 0-299 ms. and 300-800 ms were tried to segment from EEG signals. As a result, practical feedback was observed in both online and offline experiments \cite{ref67}. 

In 2014, Tsuda et. al presented BCI based visual object detection using P3 wave of 4 channels 256 Hz sampled EEG signals and oddball paradigm. As visual stimulus four different images were used. LDA, k-NN and nearest mean classifier were applied and compared with each other\cite{ref68}. 

In 2015, Bai et. al presented a hybrid BCI system, relies on both P3 wave and motor imagery of 250 Hz sampled EEG data that could operate an explorer. The system included BCI mouse, BCI speller and an explorer. The system was tested on 5 subjects with SVM classifier and promising results were achieved \cite{ref69}.

In our study, improving computationally-efficiency of pseudoinversing model in ELM was aimed.  From this point of view, theoretical analysis were performed with considering flop counts in linear algebraic operations of proposed methods. In addition empirical analysis were performed using ERP-based BCI with proposed methods. As a result of subject-dependent operations, it is reached that if priority was given to training pace, Hessenberg decomposition method, whereas if priority was given to performances Householder reflection method should be replaced with SVD. 

%In our study, visual stimuli based BCI system experiment conducted with P3 wave of ERP. In the experiment, conventional ELM and proposed novel ELM models; LuELM, HessELM, SchurELM, hhQRELM, mgsQRELM were applied to P3 wave of EEG signal to classify and detect objects. In addition, they were compared with each other in terms of several performance measurements and training duration. 19 subjects participated and the results showed that if priority was given to training pace, Hessenberg decomposition method, and if priority was given to performance measures Householder reflection method can replace SVD. In addition other proposed methods gave comparable results.

This paper is organized as follows; in Sect. \ref{sec:Materials and Methods} experimental setup, EEG date acquiring, preprocessings and classification algorithms are explained. Performance results are given in Sect. \ref{sec:Results}. The results are discussed in Sect. \ref{sec:Discussion}. Conclusions are given in Sect. \ref{sec:Conclusions}.

\section{Materials and Methods} \label{sec:Materials and Methods}
\begin{figure}
    \centering
    \includegraphics[width=0.45\textwidth]{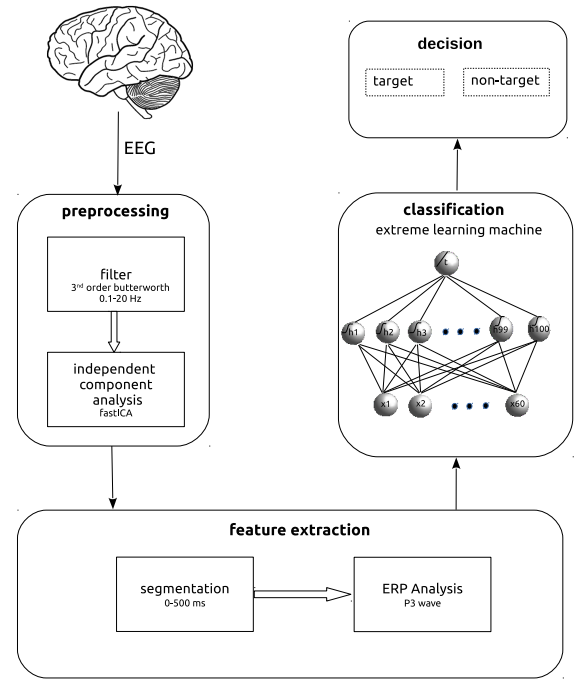}
    \caption{The proposed system for ERP based BCI}    \label{flowchart}
\end{figure}

The system architecture and data processing are illustrated in Figure \ref{flowchart}.

%\subsection{Data Acquisition and Ethics}
%\label{sec:2}Data Acquisition
%In this paper, EEG visual stimuli database, recently constructed by authors of this paper \cite{r40}, was used. It includes six number of scenarios, in this paper only the $6^{th}$ scenario of it was used for comparison of classifiers. Visual stimulus used are illustrated in Figure \ref{fig:visualobjects}.
\begin{figure}%
 \centering
    \label{fig:F}%
   \includegraphics[width=0.4\textwidth]{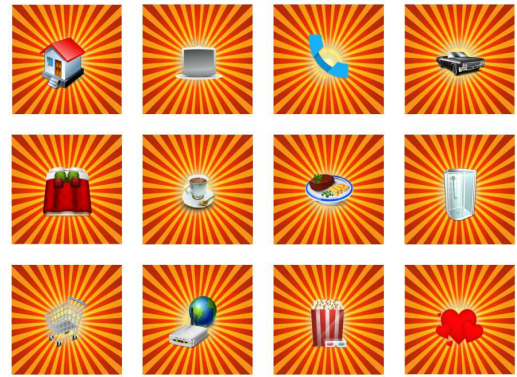}
    \caption{Visual stimulus of the scenario. (small colour image with orange concentric background)}%
   \label{fig:visualobjects}%
\end{figure} 
\subsection{Participants and Ethic} 
19 number of healthy male subjects aged between 17 and 36 (average 21.04 $\pm$2.03) participated to the experiment. None of the participants had experienced any BCI application, before. Exclusion criteria were cognitive, eyesight or hearing deficit, alcohol abuse, illiteracy, history of epilepsy or any neurological disorder. The experiment received approval from Medical School of Mustafa Kemal University and was conducted in compliance with good clinical guidelines and Helsinki declaration \cite{ref60}. All participants were informed about EEG recording procedure and asked to sign a consent form. %Ethical issues in the Declaration of Helsinki, World Medical Association $-$Ethical Principles for Medical Research Involving Human Subjects were explained before the experiment started.  
\subsection{Experimental Setup and Data Acquisition}
\begin{figure*}%
  \centering
    \includegraphics[width=1\textwidth]{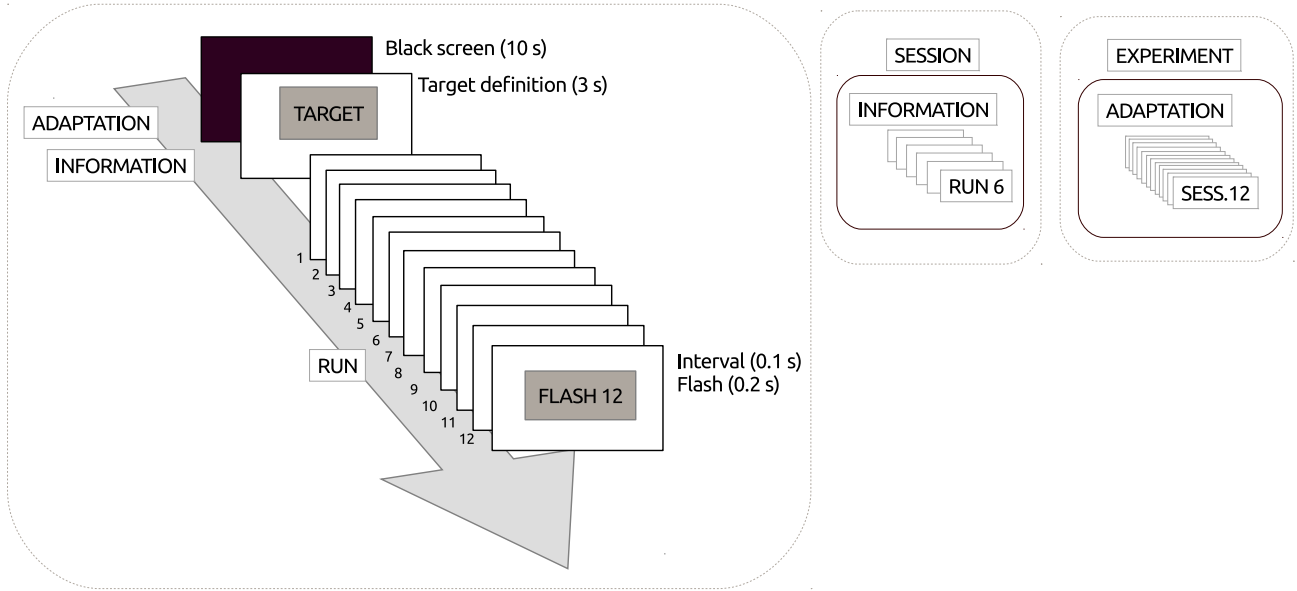}
    \caption{Experimental procedure. At the beginning of each experiment, in adaptation interval of a black screen was shown for 10 s. Following, the information interval displayed a target image for 3 s which subject was to follow. Each subject completed 12 sessions each of which consists of 6 runs. Each run comprises random flashing of 12 images (Figure \ref{fig:visualobjects}).}%
  \label{fig:exp}%
\end{figure*}

The experiment (Figure \ref{fig:exp}) was implemented in OpenVIBE software \cite{ref16}, consisted of 12 number of images (Figure \ref{fig:visualobjects}) obtained via Google Images and downsampled to a resolution of $480 \times 480$ pixels. Data recording were implemented in OpenVIBE and preprocessing were implemented in MATLAB\textregistered  with the support of the open-source EEGLAB toolbox \cite{ref18} and were performed as follows:
\begin{itemize}
\renewcommand{\labelitemi}{$\bullet$}
\item EEG data were acquired using a 14-channel Emotive EPOCH+ amplifier.
\item Electrodes were mounted according to the standard international $10-20$ system. Reference and ground electrodes were placed respectively on the mastoids and earlobes. 
\item Data were sampled at the frequency of 128 Hz. %Recordings were bandpass filtered offline using a $4^{th}$ order Butterworth filter with cut-off frequencies of 0.2 Hz and 20 Hz.
% \item Segments of 500 milliseconds from the onset of the stimulus, were extracted.
% \item Independent component analysis (fast ICA) was applied to remove ocular artifacts.
% \item Trials were inspected visually and those still affected by artifacts were discarded.
\end{itemize}

\subsection{Preprocessings and Feature Extraction}
Recordings were bandpass filtered offline using a $3^{rd}$ order Butterworth filter with cut-off frequencies of 0.2 Hz and 20 Hz. 

Independent component analysis (fast ICA) was applied to remove ocular artifacts. Trials were inspected visually and those still affected by artifacts were discarded.

\begin{figure}
    \centering
    \includegraphics[width=0.5\textwidth]{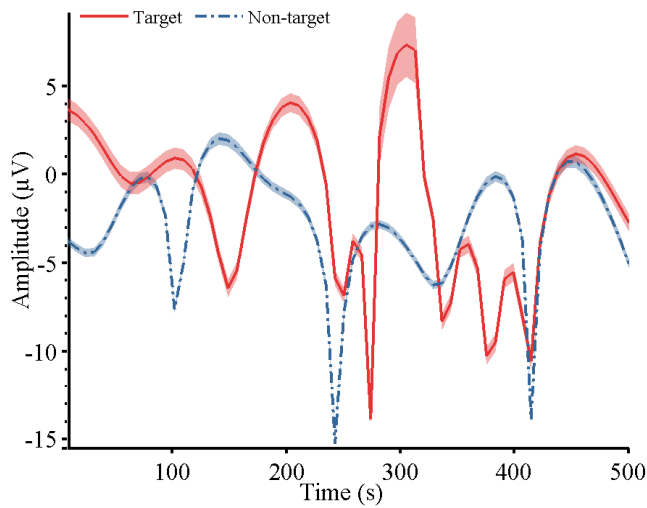}
    \caption{Grand average of segments and standard error $-$for all channels}%
    \label{fig:p300}
\end{figure}

Segments of 500 ms from onset of stimuli were extracted and grand averaged across all the channels (Figure \ref{fig:p300}) to reveal ERP components and used as features.

Prior to classification, features were linearly normalized  within the range $[0, +1]$ using Eq. (\ref{linearnormalization}),

\begin{equation}
\label{linearnormalization}
    x_{norm}=\left(\frac{x-x_{min}}{x_{max}-x_{min}}\right)
\end{equation}
where $x_{min}$ and $x_{max}$ represent respectively the lowest and highest values of each feature. The normalization coefficients, which were extracted from the training data, were stored to normalize the test data as well.

\subsection{Classification}
\subsubsection{Extreme Learning Machine (ELM)}
\begin{figure}
\centering

\tikzset{%
   neuron missing/.style={
    draw=none, 
    scale=4,
    text height=0.333cm,
    execute at begin node=\color{black}$\vdots$
  },
}

\begin{tikzpicture}[x=1.5cm, y=1.5cm, >=stealth]

\foreach \m/\l [count=\y] in {1,2,3}
{
 \node [circle,fill=gray!50,minimum size=1cm] (input-\m) at (0,2.5-\y) {};
}
\foreach \m/\l [count=\y] in {4}
{
 \node [circle,fill=gray!50,minimum size=1cm ] (input-\m) at (0,-2.5) {};
}
 
 \node [neuron missing]  at (0,-1.5) {};

\foreach \m [count=\y] in {1}
  \node [circle,fill=gray!50,minimum size=1cm ] (hidden-\m) at (2,0.75) {$\Sigma+b_1$};
  
\foreach \m [count=\y] in {2}
  \node [circle,fill=gray!50,minimum size=1cm ] (hidden-\m) at (2,-1.85) {$\Sigma+b_m$};
  
 \node [neuron missing]  at (2,-0.3) {};

\foreach \m [count=\y] in {1}
  \node [circle,fill=gray!50,minimum size=1cm ] (output-\m) at (4,0.6-\y) {$\Sigma$};
 
 %\node [neuron missing]  at (4,-0.4) {};

\foreach \l [count=\i] in {1,2,3,n}
  \draw [<-] (input-\i) -- ++(-1,0)
    node [above, midway] {$x_{\l}$};

\foreach \l [count=\i] in {1,m}
  \node [above] at (hidden-\i.north) {$\mathbf{H}_{\l}$};

\foreach \l [count=\i] in {1}
  \draw [->] (output-\i) -- ++(1,0)
    node [above, midway] {$t$};

   %input to hidden
  
    \draw [->] (input-1) -- (hidden-1)
    node[above, midway]{$v_{11}$};
    \draw [->] (input-1) -- (hidden-2)
    node at (1.1,0) {$v_{1m}$};
   
   \draw [->] (input-2) -- (hidden-1)
    node[above, midway]{$v_{21}$};
    \draw [->] (input-2) -- (hidden-2)
     node at (1,-0.4) {$v_{2m}$};
    
    \draw [->] (input-3) -- (hidden-1)
    node[above, midway]{$v_{31}$};
    \draw [->] (input-3) -- (hidden-2)
   node at (1,-1){$v_{3m}$};
    
    \draw [->] (input-4) -- (hidden-1)
    node at (0.45,-1.4) {$v_{n1}$};
    \draw [->] (input-4) -- (hidden-2)
    node at (0.6,-2.1) {$v_{nm}$};
    
    % hidden to output
    \draw [->] (hidden-1) -- (output-1)
     node[above, midway]{$w_1$};
     \draw [->] (hidden-2) -- (output-1)
     node[above, midway]{$w_m$};

\foreach \l [count=\x from 0] in {Input, Hidden, Output}
\node [align=center, above] at (\x*2,2) {\l \\ layer};

\end{tikzpicture}
\caption{Structure of ELM}
\label{Elmstructure}
\end{figure}
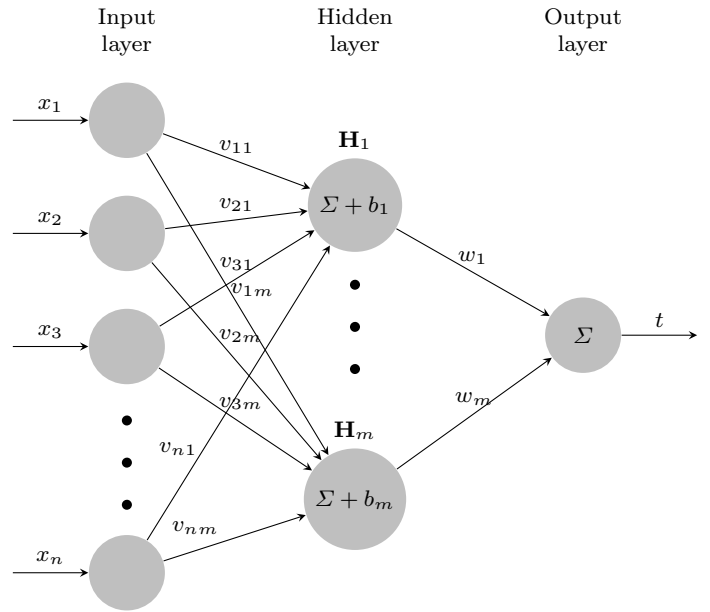

Conventional ELM \cite{ref23} is a fully-connected single-hidden layer feedforward neural network (SLFN) that has random number of nodes in hidden layer. Its architectural structure is illustrated in Figure \ref{Elmstructure}. In the input layer weights and biases are assigned randomly, whereas in the output layer weights are computed with non-iterative linear optimization technique based on generalized-inverse. Hidden layer with non-linear activation function makes non-linear input data linearly-separable.  
%In the input layer weights and biases are assigned randomly, and in the output layer weights are determined mathematically. ELM uses Moore-Penrose pseudoinverse conditions \cite{r44} with SVD (explained in Appendix \ref{svdsection}) for learning process. 

Let ${(x_{i},t_{i})}$ be a sample set, with $n$ distinct samples, where $\mathbf{x}_i=[x_{i1}, x_{i2}, \dots, x_{in}]^T$  is the $i^{th}$ input sample and ${t}_i$ is the $i^{th}$ desired output. With $m$ hidden neurons, the output of $k^{th}$ hidden layer is given by (where $k<m$);
 
\begin{equation}
\mathbf{H}_{ik}=\varphi(\sum_{i=1}^n x_{ik}{v}_{ik}+b_{k})
\label{ELM1}
\end{equation}
 and $k^{th}$ desired output is given by;
\begin{equation}
\mathbf{t_{k}}= \sum_{j=1}^mH_{jk}{w}_j 
\label{ELM2}
\end{equation}

\begin{equation}
\mathbf{H}w=t
\label{ELM3}
\end{equation}
where $\varphi(\cdot)$ is the activation function, $\mathbf{H}=[h_{i1} \dots h_{im}]$ is the output of hidden neurons, $\mathbf{v}=[v_{i1}, \dots, v_{in}]$ is the input layer weight matrix, $\mathbf{w}=[w_{1},\dots, w_{m}]^T$ is the output layer weight matrix, $\mathbf{b}=[b_1, \dots, b_m]$ is the bias value of hidden neuron and $t_{k}$ is the desired target in the training set.

Finally, the output weight matrix $w$ is calculated using; 
\begin{equation}
\mathbf{w'}=\mathbf{H}^{-1}\mathbf{t}.
\label{w1}
\end{equation}

In training set ELM with $n$ neurons in hidden layer, approximates input samples with zero error such that $\sum_{j=1}^{m}||t'_j-{t}_j||=0$ where  ${t'}_j$ is network output computed with using $w'$ in Eq.(\ref{w1}). But in this case due to overfitting generalization capacity becomes very poor. Hidden layer neuron number $m$ must be selected randomly or empirically, such that $m<n$ to prevent overfitting.

Inverse of $\mathbf{H}$ can not be determined directly if $\mathbf{H}$ is not a full-rank matrix. Pseudoinverse of the matrix $\mathbf{H}$, namely $\mathbf{H}^{+}$, can be computed via least square solution:
\begin{equation}
\mathbf{H}^{+}  = (\mathbf{H}^{T}\mathbf{H})^{-1}\mathbf{H}.
\label{w2}
\end{equation}

This solution is accurate as long as square matrix $(\mathbf{H}^{T}\mathbf{H})$ is invertible. In SLFN, it is singular in most of the cases since there is tendency to select $m<<n$. In conventional ELM, Huang et. al. \cite{ref23} has solved this problem using SVD \cite{ref26}  approach. However, SVD is very slow when dealing with large data and has low-convergence to real solution. \cite{ref33,ref34}.

\subsubsection{Lower Upper Triangularization ELM $-$LuELM}
Hidden layer output matrix $\mathbf{H}$ can be decomposed as $\mathbf{H}=LU$ where $L$ is a lower triangular matrix and $U$ is an upper triangular matrix  using (LU) triangularization. %(described in Appendix \ref{lusection}).

$\mathbf{H}w=t$ in Eq. (\ref{ELM3}) can be solved directly. The overall procedure for solving $\mathbf{H}w=t$ is explained as follows,
\begin{itemize}
\renewcommand{\labelitemi}{$\bullet$}
\item Decompose \footnote{LU decomposition works for both square and non-square matrices, and least square solution is not needed.} $\mathbf{H}$ such that $\mathbf{H}=LU$. Hence $LUw=t$
\item Let $Uw=y$, so that $Ly=t$. Solve this system using forward substitution.
\begin{equation}
\begin{split}
y_1 &=t_1/L_{1,1} \\
y_2 &=(t_2-(L_{2,1}y_1))/L_{2,2} \\
y_3 &=(t_3-(L_{3,1}y_1)+(L_{3,2}y_2))/L_{3,3} \\
\vdots \\
y_i &=(t_i-\sum_{j=1}^{i-1}L_{ij}y_j)/L_{ii}
\end{split}
\end{equation}
\item Solve the triangular system $Uw=y$ using backward substitution.
\begin{equation}
\begin{split} 
w_{e} &=y_{e}/U_{e,e} \\  
w_{e-1} &=(y_{e-1}-(U_{e-1,e}w_{e}))/U_{e-1, e-1} \\
w_{e-2} &=(y_{e-2}-(U_{e-2,e-1}w_{e-1})+(U_{e-1, e}w_{e}))/U_{e-2,e-2} \\
\vdots \\
w_i &=(y_i-\sum_{j=i+1}^nU_{ij}w_j)/U_{ii}
\end{split}
\end{equation}
\end{itemize} 
\subsubsection{Modified Gram Schmidt Process QR Decomposition ELM $-$MgsQRELM}
Using MGS process, hidden layer output matrix $\mathbf{H}$ can be decomposed as\footnote{Modified Gram Schmidt algorithms works for both square and non-square matrices, and least square solution is not needed.},
\begin{equation}
\label{mgseq1}
\mathbf{H}=QR
\end{equation}
where $Q$ is an orthogonal matrix that has orthonormal columns and rows and $R$ is an upper triangular matrix. Inverse of matrix $\mathbf{H}$ is,
\begin{equation}
\label{mgseq2}
\mathbf{H}^{-1}=R^{-1}Q^{T}
\end{equation}
Determining accurate inverse of upper triangular matrix $R$ is important. Determinant of any triangular matrix can be computed as product of its main diagonal elements \cite{r39}. Since it is clear that upper triangular matrix $R$ has non-zero main diagonal elements, determinant of $R$ is always non-zero. Therefore, when considering singularity conditions, it is reached that $\begin{vmatrix} R \end{vmatrix} \neq 0$  and matrix $R$ is non-singular so,  $R^{-1}$ exists.
Target values are reached and learning is achieved as follows,
\begin{itemize}
\renewcommand{\labelitemi}{$\bullet$}
\item In Eq. (\ref{w1}) $\mathbf{H}^{-1}$ is substituted in Eq. (\ref{mgseq2})  and output weights $w$ are reached.
\item In Eq. (\ref{ELM3}), $w$ is put in its place.
\end{itemize} 
\subsubsection{Schur Decomposition ELM $-$SchurELM} 
Since Eq. (\ref{ELM3}) is an underdetermined system of equation, using Eq. (\ref{w2}) pseudoinverse of hidden layer output matrix $\mathbf{H}$ is formed as\footnote{Schur decomposition works only for square matrices, therefore least square solution is needed.},
\begin{equation}
\label{schurelmeq1}
\mathbf{H}^{+}=\mathbf{H}^T(\mathbf{H}^T\mathbf{H})^{-1}
\end{equation}
To reach $\mathbf{H}^{+}$, square matrix $\mathbf{H}^T\mathbf{H}$ can be decomposed using Schur decomposition,
\begin{equation}
\label{schurelmeq2}
\mathbf{H}^T\mathbf{H}=UTU^{*}
\end{equation}
where $U$  is a unitary matrix and $T$  is an upper triangular matrix. When $\mathbf{H}^T\mathbf{H}$ is substituted in Eq. (\ref{schurelmeq1}),
\begin{equation}
\label{schurelmeq3}
\begin{split}
\mathbf{H}^{+} &=\mathbf{H}^T(UTU^{*})^{-1} \\
&=\mathbf{H}^TUT^{-1}U^{*}
\end{split}
\end{equation}
where $T$ is an upper triangular matrix that has non-zero elements in its main diagonal. Therefore,$\begin{vmatrix} U \end{vmatrix} \neq 0$  and $U^{-1}$ exists.
Target values are reached and learning is achieved as follows,
\begin{itemize}
\renewcommand{\labelitemi}{$\bullet$}
\item In Eq. (\ref{w1}) $\mathbf{H}^{-1}$ is substituted with $\mathbf{H}^{+}$ in Eq. (\ref{schurelmeq3}) and output weights $w$ are reached.
\item In Eq. (\ref{ELM3}), $w$ is put in its place.
\end{itemize} 
\subsubsection{Householder Reflection QR Decomposition $-$HhQRELM} 
Using householder reflection, hidden layer output matrix $\mathbf{H}$ can be decomposed as\footnote{Householder reflection algorithm works for both square and non-square matrices, and least square solution is not needed.},
\begin{equation}
\label{hhQRELMeq1}
\mathbf{H}=QR
\end{equation}
where $Q$ is an orthogonal matrix that has orthonormal columns and rows, $R$ is an upper triangular matrix. Inverse of $\mathbf{H}$ is,
\begin{equation}
\label{hhQRELMeq2}
\mathbf{H}^{-1}=R^{-1}Q^T
\end{equation}
Determining accurate inverse of upper triangular matrix $R$ is important. Determinant of any triangular matrix can be computed as product of its main diagonal elements \cite{r39}. Since it is clear that upper triangular matrix $R$ has non-zero main diagonal elements, determinant of $R$ is always non-zero. Therefore, when considering singularity conditions, it is reached that $\begin{vmatrix} R \end{vmatrix} \neq 0$  and matrix $R$ is non-singular, therefore $R^{-1}$ exists.
Target values are reached and learning is achieved as follows,
\begin{itemize}
\renewcommand{\labelitemi}{$\bullet$}
\item In Eq. (\ref{w1}) $\mathbf{H}^{-1}$ is substituted in Eq. (\ref{hhQRELMeq2})  and output weights $w$ are reached.
\item In Eq. (\ref{ELM3}), $w$ is put in its place.
\end{itemize} 
\subsubsection{Hessenberg Decomposition ELM $-$HessELM} 
%\label{HessELMsection}
Since Eq. (\ref{ELM3})  is an underdetermined system of equation, using Eq. (\ref{w2}) pseudoinverse of hidden layer output matrix $\mathbf{H}$ is formed as\footnote{Hessenberg decomposition works only for square matrices, therefore least square solution is needed.},
\begin{equation}
\label{hesselmeq1}
\mathbf{H}^{+}=\mathbf{H}^T(\mathbf{H}^T\mathbf{H})^{-1}
\end{equation}
To reach $\mathbf{H}^{+}$, square matrix $\mathbf{H}^T\mathbf{H}$ can be decomposed using hessenberg decomposition %(described in Appendix \ref{hessdecompositionsection}),
\begin{equation}
\label{hesselmeq2}
\mathbf{H}^T\mathbf{H}=QUQ^{*}
\end{equation}
where $Q$  is a unitary matrix and $U$  is an upper hessenberg matrix. When $\mathbf{H}^T\mathbf{H}$ is substituted in Eq. (\ref{hesselmeq1}),
\begin{equation}
\label{hesselmeq3}
\begin{split}
\mathbf{H}^{+} &=\mathbf{H}^T(QUQ^{*})^{-1} \\
&=\mathbf{H}^TQU^{-1}Q^{*}
\end{split}
\end{equation}
where $U$ is an upper hessenberg matrix that is also symmetric and tridiagonal. When considering singularity conditions, it is reached that $\begin{vmatrix} U \end{vmatrix} \neq 0$  and matrix $U$ is non-singular therefore, $U^{-1}$ exists.
Target values are reached and learning is achieved as follows,
\begin{itemize}
\renewcommand{\labelitemi}{$\bullet$}
\item In Eq. (\ref{w1}) $\mathbf{H}^{-1}$ is substituted with $\mathbf{H}^{+}$ in Eq. (\ref{hesselmeq3}) and output weights $w$ are reached.
\item In Eq. (\ref{ELM3}), $w$ is put in its place.
\end{itemize}

\subsection{Complexity Analysis}
Complexity analysis is performed to decide which method is the most useful and relevant, in theory. Complexity can be defined in terms of flops required to reach desired solution \cite{ref61}. Flop that represents a multiplication, division, addition or subtraction, is accepted as a basic unit of computation. Number of required flops in each matrix decomposition for any given $m \times n$ matrix are given in Table \ref{table:flopcounts}.

\begin{table}
\centering
\caption{Flop counts of methods ($n>m$)}
\label{table:flopcounts}
\begin{tabular}{@{}ll@{}}
\toprule
Method                                                                               & Flop Count \\ \midrule
\begin{tabular}[c]{@{}l@{}} Householder Reflection\end{tabular} &   $2n^2(m-\frac{n}{3})$         \\
\begin{tabular}[c]{@{}l@{}}Modified-Gram Schmidt \end{tabular}   &           $2mn^2$ \\
Singular Value Decomposition                                                       &     $2mn+11n^3$         \\
LU Triangularization                                                                &        $\frac{2n^3}{3}$     \\
Hessenberg Decomposition                                                           &      $\frac{10n^3}{3}$        \\
Schur Decomposition                                                                &      $2mn^2$        \\ \bottomrule
\end{tabular}
\end{table}

Singular value decomposition is very expensive in computation, (see Table \ref{table:flopcounts}) whereas other approaches are relatively much cost-effective.  

\subsection{Statistical Methods}
To validate effects of classifiers on performance rates, multivariate analyze of variant (MANOVA) test was applied on IBM SPSS software platform. MANOVA test was preferred because there were more than one independent variable in dataset. MANOVA test indicates that whether a meaningful difference between classifiers and means of performances rates exist or not. Dependent variable and independent variable are given below:
Dependent variable: Classifier. It includes information of related classifier 1, 2, 3, 4, 5, 6, 7, 8 and 9.
Independent variables: Sensitivity, precision, f measure, specificity, MCC and accuracy performance measures.

Hypothesis to compare mean values are given below (p significance value: 0.05):

$H_0$ : There is no statistical meaningful difference between classifier and sensitivity, precision, f measure, specificity, MCC and accuracy parameters of classifiers with 95\% confidence.

$H_1$ : There is a statistical meaningful difference between classifier and sensitivity, precision, f measure, specificity, MCC and accuracy parameters of classifiers with 95\% confidence.

\section{Results}
\label{sec:Results}

In our study, five number of proposed learning approaches were applied to P3 component of ERP to detect visual objects on a subject-dependent basis. Additionally, the results of them were compared with those of conventional machine learning algorithms $-$support vector machine (SVM), multi-layer perceptron (MLP) and k-nearest neighbour k-NN classifiers. SVM used radial-basis kernel, k-NN measured euclidean distance and k=5, MLP with gradient descent algorithm had three number of computational hidden layers with 30, 20 and 10 neurons each of which had logistic sigmoid activation function. Learning rate of $0.1$ was fixed at each neuron.

%Proposed approaches were compared with each other in terms of several performance measurements (sensitivity, precision, specificity, f-measure, matthews correlation coefficient (MCC), overall accuracy) and training duration. 

%In classifying process of ELMs, feature data were divided into train and test parts. At first train part was tried to be classified. Unless train part can be classified accurately, test part can no longer be classified. Also if train part is classified completely (almost zero error), due to overfitting test part cannot be classified. To balance these circumstances optimum $k$ hidden layer neuron must be determined ($k<m$). After experimental analysis 100 number of neurons were preferred in hidden layer of both ELM and its proposed versions. In ELM and its proposed versions logistic sigmoid transfer function was used. 

%In preprocessing part, to reveal P3 wave, segments of 500 ms duration after stimuli onset were extracted, butterworh band pass filtered between 0.1 Hz and 20 Hz in $3^{rd}$ order and fast ICA was applied to discard EEG components that are based on muscle and ocular artifacts. 

Randomly dividing into train and test part, like in conventional cross-validation technique, definitely results in problems, for database used in this study. Data should be divided regularly according to $6$ runs for each image that has $72$ features. Accordingly, each time every $6$ runs for each image were used as testing part when residuals ($12-1=11$) as training part. In brief, in this study $12$-fold cross-validation technique that determines testing and training parts systematically was designed to achieve validated performance results.

Database used in this study has unbalanced distributed classes ($72$ target and $792$ non-target). When classifier incorrectly classifies all minority classes 91.6\% overall accuracy occurs, although there is no learning. Therefore, performance measurements that take into account both minority and majority classes help us to observe whether learning exists, or not. From this point of view; sensitivity, precision, specificity, matthews correlation coefficient (MCC) performances measurements were obtained from confusion matrices.

\begin{table*}
\centering
\caption{Average values of performance measurements for each classifier}
\label{table1}
\begin{tabular}{@{}lcccccccc@{}}
\toprule
\multirow{2}{*}{Classifer} & \multirow{2}{*}{\begin{tabular}[c]{@{}c@{}}Sensitivity\\ (\%)\end{tabular}} & \multirow{2}{*}{\begin{tabular}[c]{@{}c@{}}Precision\\ (\%)\end{tabular}} & \multirow{2}{*}{\begin{tabular}[c]{@{}c@{}}F measure\\ (\%)\end{tabular}} & \multirow{2}{*}{\begin{tabular}[c]{@{}c@{}}Specifity\\ (\%)\end{tabular}} & \multirow{2}{*}{\begin{tabular}[c]{@{}c@{}}MCC\\ (\%)\end{tabular}} & \multirow{2}{*}{\begin{tabular}[c]{@{}c@{}}Accuracy\\ (\%)\end{tabular}} & \multicolumn{2}{c}{Duration (s)} \\ \cmidrule(l){8-9} 
                           &                                                                             &                                                                           &                                                                           &                                                                           &                                                                     &                                                                          & Train           & Test           \\ \cmidrule(r){1-9}
ELM                        & 99,70                                                                       & 98,03                                                                     & 98,74                                                                     & 99,82                                                                     & 98,71                                                               & 99,81                                                                    & 0,06            & 0,001          \\
HessELM                    & 99,55                                                                       & 98,25                                                                     & 98,80                                                                     & 99,88                                                                     & 98,75                                                               & 99,81                                                                    & 0,04            & 0,001          \\
LuELM                      & 99,39                                                                       & 98,25                                                                     & 98,70                                                                     & 99,84                                                                     & 98,66                                                               & 99,79                                                                    & 0,05            & 0,001          \\
HhQRELM                    & 99,61                                                                       & 98,68                                                                     & 99,05                                                                     & 99,88                                                                     & 99,03                                                               & 99,85                                                                    & 0,05            & 0,001          \\
MgsQRELM                   & 99,61                                                                       & 98,68                                                                     & 99,05                                                                     & 99,88                                                                     & 99,03                                                               & 99,85                                                                    & 0,11            & 0,001          \\
SchurELM                   & 99,63                                                                       & 98,46                                                                     & 98,95                                                                     & 99,86                                                                     & 98,92                                                               & 99,84                                                                    & 0,05            & 0,001          \\
SVM                        & 94.80                                                                        & 95.50                                                                      & 96.70                                                                      & 95.60                                                                      & 94.50                                                                & 97,60                                                                    & 0,50            & 0.020           \\
MLP                        & 93,70                                                                       & 95,80                                                                     & 94,70                                                                     & 95,70                                                                     & 94,60                                                               & 95,30                                                                    & 1,10            & 0.300            \\
k-NN                       & 99,56                                                                       & 98,46                                                                     & 98,91                                                                     & 99,87                                                                     & 98,88                                                               & 99,46                                                                    & 0,13            & 0.020           \\ \bottomrule
\end{tabular}
\end{table*}

\begin{figure}
\centering
\begin{tikzpicture}
\begin{axis}[
    width=3.2in,
    height=2in,
    axis x line=center,
    axis y line=left,
    symbolic x  coords={HessELM,ELM,LuELM,MgsQRELM,HhQRELM,SchurELM},
    enlargelimits=true,
    ymin=0,
    ymax=19,
    nodes near coords,
    ylabel style={align=right},
    ylabel={Number of subjects},
    x tick label style={font=\small,text width=2cm,align=left}, ybar, tick label style={rotate=90}
    ]
    \addplot[color=blue,fill] coordinates {(MgsQRELM,12) (HhQRELM,12)};  
    \addplot[color=black,shade] coordinates {(HessELM,9) (ELM,10) (LuELM,10) (SchurELM,10)};
\end{axis}
\end{tikzpicture}
\caption{Number of subject-based classifications that had 100\% performance in all measurements}
\label{fig:fullacc}
\end{figure}
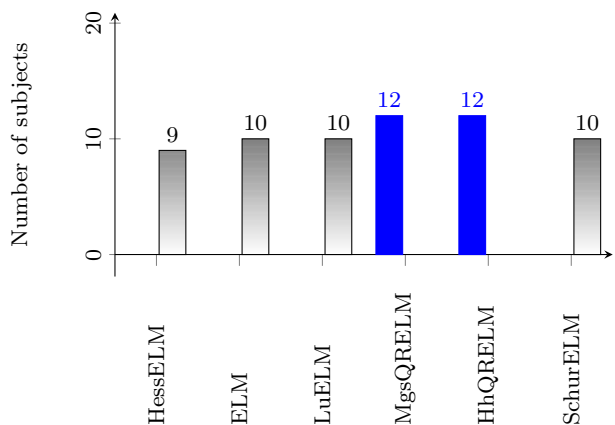

%Using each classifier for each individual subject, subject-based classification is performed. Hence, designed BCI system can be used by any subject whose data were trained and generalized before. 
Average performance measurements for training and test parts are given in Table \ref{table1}. For each classifier, number of subject-based test data classifications reached performance of 100\% in aforementioned measures are shown in Figure \ref{fig:fullacc}.

Results of MANOVA test including  Pillai's Trace, Wilks' Lambda, Hoteling's Trace and Roy's Largest Root are given in Table \ref{manovaresult}. Additionally, results of variances homogeneity test (Levene's test of equality of error variances) are given in Table \ref{homoresult}. 

\begin{table}
\centering
\caption{Results of MANOVA test}
\label{manovaresult}
\begin{tabular}{@{}lccccc@{}}
\toprule
Effect             & Value & F     & H. df & E. df & Sig. \\ \midrule
Pillai's T.     & 0,38 & 1,37 & 48        & 966  & 0,00 \\
Wilks' L.      & 0,65 & 1,45 & 48         & 771  & 0,00 \\
Hoteling's T.   & 0,47 & 1,53 & 48        & 926  & 0,00 \\
Roy's L. R. & 0,34 & 7,00 & 8         & 161  & 0,00 \\ \bottomrule
\end{tabular}
\end{table}

\begin{table}
\centering
\caption{Results of homogeneity test.}
\label{homoresult}
\begin{tabular}{@{}lcccc@{}}
\toprule
Independent Variables & F     & df1 & df2 & Sig. \\ \midrule
Sensitivity           & 0,965 & 8   & 161 & 0,46 \\
Precision             & 6,808 & 8   & 161 & 0,00 \\
F measure             & 5,353 & 8   & 161 & 0,00 \\
Specificity,          & 5,015 & 8   & 161 & 0,00 \\
MCC                   & 5,334 & 8   & 161 & 0,00 \\
Accuracy              & 4,463 & 8   & 161 & 0,00 \\ \bottomrule
\end{tabular}
\end{table}

\section{Discussion}
\label{sec:Discussion}
In this paper, 5 number of methods to improve computational stability of ELM were proposed. They are relatively much cost-effective than ELM, in theory (see Table \ref{table:flopcounts}).They were applied to ERP-based BCI to compare with each other and conventional machine learning algorithms empirically. 

Averaging time series of 500 ms length from stimuli onset in ERP analysis aimed at describing the P300 component across all EEG channels. As suggested by the average time-locked EEG data (Figure \ref{fig:p300}), it appears to be a difference within range of 280 and 340 ms, with segments of target trials recording higher values than non-target ones. This tendency might be better captured by the utilized averaging features, which works as effective predictors.

Generally, two main criteria should be considered to decide which classifier is the most useful and efficient; being fast in both training and testing, and accurate in decision making. 

Because BCI system was not applied in an adaptive way, test duration were much more important in practice. For ELMs, in the course of testing forward pass is implemented with weights computed in training. The expectancy is retrieving almost equal testing duration, because they have completely same architecture (Table \ref{table1}). However testing duration of SVM, MLP and k-nn are very long when compared to ELMs. In an attempt to compare usefulness and efficiency of the methods, in this study training duration were also considered. 

In Table \ref{table1}, although classifiers are seen to have almost the same performances, one should note that any miniature increment over 91.6\% means highly improvement in learning capability. When considering Table \ref{table1}, it is obviously seen that not only training duration of HessELM is shorter than that of conventional ELM, but also almost all performance measurements of HessELM are relatively higher than that of conventional ELM.

Except for MgsQRELM, training speed of proposed learning approaches are higher than that of conventional ELM. This proves that this paper contributes implementation of fast learning approaches. Though, HhQRELM and MgsQRELM have same performance rates in all performance measures, they apparently differ in their training speed. As it is seen training speed of HhQRELM is 2 times faster than that of MgsQRELM. Therefore, when MGS process and householder reflection methods are compared, householder reflection is more effective than MGS process. 

Conventional ELM has the highest performance rate only in sensitivity measure and its training speed is apparently lower than others, except for MgsQRELM. Though, LuELM is as fast as HhQRELM in training process, it has lower performance rates than HhQRELM in all performance measures. When LuELM and SchurELM are compared, although they have same training speed, SchurELM has higher performance rates than LuELM in all performance measures.

Training pace of HessELM classifier is faster than that of other classifiers. In detail, training pace of HessELM is almost $1.5$ times faster than that of ELM, almost $2.75$ times faster than that of MgsQRELM, almost $1.25$ times faster than that of LuELM and HhQRELM and SchurELM, almost 27 times faster than that of MLP, almost 12 times faster than that of SVM and almost 2.5 times faster than that of k-NN classifier. 

In machine learning systems reaching short training duration is of utmost importance. Therefore proposed HessELM classifier serves as a more effective classifier than others. 

In Figure \ref{fig:fullacc} it is seen that almost more than half of subject-based classification reached 100\% performance rate. In Figure \ref{fig:fullacc}, it is seen that 12 of 19 subjects' subject-based classification performance measures are 100\% in all performance measures when MgsQRELM and HhQRELM classifiers are used. However, in Table \ref{table1} it is seen that MgsQRELM is the slowest in training. When HhQRELM and HessELM are compared, though HhQRELM has higher performance rates and provides more subjects to get 100\% performance rates, HessELM has faster training pace than HhQRELM.

Significance values of tests in Table \ref{manovaresult} are smaller than p significance value. In this case $H_1$ is valid. Therefore, there is a statistical meaningful difference between classifiers and sensitivity, precision, f measure, specificity, MCC and accuracy rates with 95\% confidence.

In Table \ref{homoresult}, it is seen that homogeneity of variances does not exist. And to decide which classifier has most effect on performances, Tamhane's T2 method was applied for multiple comparisons (Because homogeneity of variances does not exist, Tamhane's T2 is preferred). Tamhane's T2 indicated that HessELM classifier has more meaningful mean difference when compared to others. 

Mean difference parts of Tamhane's T2 prove that HessELM classifier has statistical meaningful difference from others. In this case, it can be said that hhQRELM classifier has statistically meaningful effect on performance rates in visual stimuli optimization problem.

Techniques in \cite{ref72} can be extended to proposed methods in this study after MSE is numerically stabilized and is re-designed to consider the convergence of zero at below and above the diagonal of $\mathbf{H}\mathbf{H}^+$. When either modified Hessenberg decomposition or Schur decomposition is used,
\begin{equation}
\begin{split}
    HAT_i &=\mathbf{H}(\mathbf{H}^T\mathbf{H}+\lambda I)^{-1}\mathbf{H}^T \\
    &=\mathbf{H}(QUQ^*+\lambda I)^{-1}\mathbf{H}^T
    \end{split}
\end{equation}
When LU triangularization is used,
\begin{equation}
\begin{split}
    HAT_i &=\mathbf{H}(\mathbf{H}+\lambda I)^{-1} \\
    &=\mathbf{H}(LU+\lambda I)^{-1}
    \end{split}
\end{equation}
When either modified Gram-Schmidt method or Householder reflection method is used,
\begin{equation}
\begin{split}
    HAT_i &=\mathbf{H}(\mathbf{H}+\lambda I)^{-1} \\
    &=\mathbf{H}(QR+\lambda I)^{-1}
    \end{split}
\end{equation}
where $HAT_i$ is $1-$diag($\mathbf{H}\mathbf{H}^+$).\\

%As a result, it can be said that if BCI system relies on ERP component that requires short feature segmentation (<1000 ms), like in this paper, pace of classifier is of utmost importance. In this case preferring HessELM classifier is the optimum choice. But if BCI system relies on ERP component that do not require short feature segmentation, instead of training pace, performance measures are of utmost importance. In this case preferring HhQRELM is the optimum choice. 

Techniques for RVFL net are needed to be improved to gain reliable solutions. Direct-input output connection can be good idea only when input data is linearly separable. In real life, mostly input data is non-linearly distributed and this always makes converging to optimum solution much harder. To avoid this and make input data linearly separable, input data should be transformed from a non-linear activation function \cite{ref29}.  RVFL net ignores this significant basic problem, whereas ELM algorithms do not. Schimdt et al \cite{refx1} did not generalize range of weights generated in hidden layer to avoid possible saturation, just specific condition for logistic sigmoid function was mentioned. Pao et al \cite{refx2} did not deal with the pseudoinversion process in the case of rank deficiency.
%Verma and Mulawka [] proposed randomized neural network with activation function at output layer. ELM never uses any activation function in output layer because it makes system over-constrained  and makes unnecessary computation.
Opposed to Guo et al \cite{refx4}, previous study \cite{ref59} approved that setting number of hidden nodes equal to number of input nodes makes system overfitted and decreases testing accuracy. In other words, it leads increasing in training, but generalization ability is definitely lost (see Figure 4 in \cite{ref59}). 

Our previous study \cite{ref59} together with current research solves difficulties mentioned in Zhang and Suganthan \cite{refx7}. Zhang and Suganthan \cite{refx11, refx12} did not suggest how to implement pseudoinverse method that complies with Moore-Pensore conditions and how to optimize regularization parameter for reliable convergence to real solution. In our current study 5 number of efficient methods were suggested for PIL and with inspirations from study \cite{ref72} regularization extensions to Hessenberg and Schur decomposition methods were enhanced.

\section{Conclusions}
\label{sec:Conclusions}
This paper introduced extensions to improve learning capacity and decrease training duration of conventional ELM. The results imply that when priority is given to training pace HessELM whereas when priority is given to performance measures hhQRELM can reach better results than conventional ELM.

%Key point in this paper is the situation of hidden layer output matrix $\mathbf{H}$. It has tendency to be ill-conditioned. If any approach that can make it full well-conditioned with computations, dot product with weight, sum with bias values or with activation function, direct solution or least square solution can converge to real solution.

Additionally, with proposed methods BCI system based on P3 wave lets severely disabled people to convey their needs and to communicate with other people easily and effectively.

As expected, close resemblance of ELM and RVFL net in development process were observed. We believe that significant improvements of each one should be incorporated to contribute in this field.

\begin{acknowledgements} 
The authors wholeheartedly thank Associate Professor Serdar Y{\i}ld{\i}r{\i}m and Associate Professor Esen Y{\i}ld{\i}r{\i}m for providing deep inspiration about optimization and neuroscience.
\end{acknowledgements}
\section*{\small{Compliance with ethical standards}}
\textbf{\small{Conflict of interest}} \small{The authors declare that there is no conflict of interest.} \\ \\
\textbf{\small{Ethical approval}} \small{All procedures performed in the current study which involved human participants were in accordance with the ethical standards of the institutional and/or national research committee and with the 1964 Helsinki declaration and its later amendments or comparable ethical standards.}

\bibliographystyle{spmpsci}      % mathematics and physical sciences
%\bibliography{data}

\end{document}